\documentclass[sigconf]{acmart}

\usepackage{booktabs}
\usepackage{amsmath}
\usepackage{graphicx}
\usepackage{placeins}
\usepackage{tikz}
\usetikzlibrary{positioning,arrows.meta,calc,shadows,decorations.pathreplacing}
\definecolor{peachLight}{HTML}{FFEDD6}    
\definecolor{peachMid}{HTML}{FFD7B0}      
\definecolor{peachDark}{HTML}{FFBC85}     
\definecolor{adapterPeach}{HTML}{FFD7B0}
\definecolor{maskGreen}{HTML}{C0DCA8}
\definecolor{embGreen}{HTML}{C8DCB8}      

\copyrightyear{2026}
\acmYear{2026}
\setcopyright{cc}
\setcctype{by}
\acmConference[RecSys '26]{20th ACM Conference on Recommender Systems}{September 27-October 02, 2026}{Minneapolis, MN, USA}
\acmBooktitle{20th ACM Conference on Recommender Systems (RecSys '26), September 27-October 02, 2026, Minneapolis, MN, USA}
\acmDOI{10.1145/3773078.3831931}
\acmISBN{979-8-4007-2284-4/2026/09}

\begin{document}

\title{Building a User Foundation Model for the Open Web}

\author{Solal Vernier}
\authornote{Both authors contributed equally to this research.}
\email{solal.vernier@teads.com}
\affiliation{%
  \institution{Teads}
  \city{Montpellier}
  \country{France}
}

\author{Ivan Can Arisoy}
\authornotemark[1]
\email{ivan-can.arisoy@teads.com}
\affiliation{%
  \institution{Teads}
  \city{Montpellier}
  \country{France}
}

\author{Merwan Barlier}
\email{merwan.barlier@teads.com}
\affiliation{%
  \institution{Teads}
  \city{Paris}
  \country{France}
}

\author{Bla\v{z} \v{S}krlj}
\email{blaz.skrlj@teads.com}
\affiliation{%
  \institution{Teads}
  \city{Ljubljana}
  \country{Slovenia}
}

\renewcommand{\shortauthors}{Vernier, Arisoy, Barlier and \v{S}krlj}

\begin{abstract}
User foundation models have demonstrated strong results in e-commerce and social recommendation, but most industrial deployments assume environments where user identity is stable and persistent. Open-web real-time bidding (RTB) operates on a structurally different data distribution: user identity is fragmented and non-persistent across browsing sessions, and the availability of browsing history depends on user privacy choices. Consequently, a significant portion of traffic carries no historical data, and available records often consist of relatively short, disjointed sessions. As a result, historical signals in this domain are typically represented as aggregated counters and recency buckets, leaving the sequential structure unexploited. To address this limitation, we present a user foundation model that applies self-supervised learning on user browsing histories and show that the learned representation improves multiple downstream production tasks, demonstrating the viability of this approach on the open web. We pre-train a Transformer encoder with masked language modeling and a sequence-level contrastive objective, then fine-tune it on the click prediction task. We optimize the encoder's pre-training pipeline with an LLM-in-the-loop search over a curated catalog of reviewable, code-level edits (lifters), instantiating the LLM-as-optimizer paradigm in an industrial setting. The same encoder representation yields +1.197\% RIG on the production bid win-rate model and +1.354\% RIG on the production CTR ranker; a 7-day live A/B test confirms +2.13\% CTR, -1.13\% eCPC (80\% CI excluding zero on both metrics).
\end{abstract}



\maketitle

\section{Introduction}
\label{sec:intro}

Foundation models for user representation have become a standard ingredient in industrial recommender systems. The pre-train-then-specialize paradigm has demonstrated consistent gains in e-commerce~\cite{shin2021one4all}, social-discovery feeds~\cite{pancha2022pinnerformer,chen2025pinfm}, mobile platforms~\cite{yang2023lurm}, content recommendation~\cite{chen2024hllm}, and large-scale ads ranking~\cite{liang2025elfm,song2026mtfm}. In most cases, a Transformer encoder pre-trained on user action sequences captures behavioral structure that supervised models trained on task labels alone tend not to recover. Every one of these systems, however, is built on a platform where user identity is stable, persistent, and attributable: a logged-in account with long, contiguous, recoverable histories.

Open-web real-time bidding (RTB) operates on a structurally different distribution. On the open web, user identity is inherently fragmented and non-persistent: a single person generates multiple disconnected browsing records across sessions, none attributable across devices or browsers. Browsing history is available only for users who have opted in and a significant part of bid requests carry no historical data at all. For users with history, available records are relatively short and bounded, limited to a window of recent sessions rather than the long, uninterrupted histories the systems described above are built on.

Within this setting, deep learning models for RTB have almost always relied on discrete feature representations. Each bid request is reduced to a fixed-size vector of hand-crafted signals (publisher, advertiser, device, context, recency and frequency buckets), and improvement has been pursued along two axes: augmenting the feature set through engineering, and advancing the ranker architecture. Both axes operate within the same representational space. The temporal ordering, sequential co-occurrence, and behavioral trajectory of user actions are not expressible within this paradigm, and the open web has historically made sequential modeling appear intractable.

We ask whether self-supervised learning on user browsing sequences is viable in this environment, and whether it can open a new, decoupled improvement axis for the production models that operate within it. Two structural constraints bound the answer. \textbf{C1~(latency):} predictions must be served at sub-millisecond latency on the bid path, shared with an existing production ranker that already consumes most of the latency budget. \textbf{C2~(sparse history):} a significant part of inference traffic has no behavioral history; opt-in users have relatively short, bounded browsing histories rather than full user trajectories. Any deployed model must produce stable, useful output across both regimes.

We answer both questions affirmatively and present a user foundation model (UFM) developed and deployed at Teads. Our contributions are:
\begin{itemize}
  \item A user foundation model pre-trained on user history sequences with masked language modeling and a sequence-level contrastive objective, fine-tuned for click prediction, and deployed under open-web RTB latency and sparse-history constraints.
  \item Offline RIG improvements from optimized encoder representations across both history regimes ($+0.99\%$ on the history-absent regime and $+1.62\%$ on the history-present regime) and with consistent gains across ranker architectures ($+1.354\%$ on GDCN~\cite{wang2023gdcn}, $+0.894\%$ on DCN$^2$~\cite{wang2025dcn2}).
  \item The same encoder representations, fine-tuned on click prediction, yield $+1.197\%$ RIG on the bid win-rate model, a structurally different prediction target, showing that the learned representation generalizes across tasks.
  \item A literature-anchored neural architecture search (NAS) workflow demonstrating that the axis has systematic headroom: encoder optimization translates into downstream gains independently of ranker changes.
\end{itemize}

\section{Related Work}
\label{sec:related}

\paragraph{Foundation models for recommender systems.} The pre-training-then-specialize paradigm has been adapted to industrial recommender systems along multiple axes. ShopperBERT~\cite{shin2021one4all} pre-trains a single Transformer encoder over e-commerce shopping behavior to produce user representations transferable across multiple downstream tasks. PinnerFormer~\cite{pancha2022pinnerformer} replaces next-action prediction with a dense all-action loss to bridge batch user-embedding generation and realtime serving, and PinFM~\cite{chen2025pinfm} scales the same paradigm to billion-scale visual discovery with deduplicated cross-attention for serving efficiency. HLLM~\cite{chen2024hllm} factors item and user modeling into a hierarchical large-language-model architecture. In online ads ranking, ExFM~\cite{liang2025elfm} serves trillion-parameter foundation models through external distillation: a 1-to-N teacher FM supplies offline-logged supervision to multiple vertical models (VMs) via a Data Augmentation Service, with Auxiliary Head and Student Adapter components addressing cross-domain bias and FM/VM freshness gaps; and MTFM~\cite{song2026mtfm} uses heterogeneous tokenization to span multiple recommendation scenarios with a single encoder. Every one of these systems assumes a platform where user identity is stable and persistent: a logged-in account with a contiguous, recoverable history. None has been designed for or validated on an open-web distribution.

\paragraph{Sequential user modeling.} Self-supervised pre-training of Transformer encoders on user histories has a long lineage: SASRec~\cite{kang2018sasrec} introduces the causal Transformer for sequential recommendation; BERT4Rec~\cite{sun2019bert4rec} adapts masked language modeling to user sequences; subsequent work extends the recipe with contrastive auxiliaries (CL4SRec~\cite{xie2022cl4srec}, DuoRec~\cite{qiu2022duorec}, $\text{S}^3$-Rec~\cite{zhou2020s3rec}, ContraRec~\cite{wang2023contrarec}) or with sequence-level dropout-based augmentations~\cite{gao2021simcse}. We adopt the BERT4Rec / MLM lineage and add a sequence-level NT-Xent objective in the spirit of CL4SRec and DuoRec, with two differences: (i) the contrastive views are obtained by a temporally-disjoint chronological partition of the user history, motivated by the bounded, fragmentary nature of open-web browsing sequences rather than CL4SRec's crop/mask/reorder augmentations or DuoRec's dropout-based positive construction; and (ii) the encoded units are interaction \emph{triplets} rather than single items, an inductive bias reflecting the impression-level structure of industrial CTR data~\cite{chen2019bst,zhou2018din}. Continuous time encodings of inter-event delay---TiSASRec~\cite{li2020tisasrec}, Time2Vec~\cite{kazemi2019time2vec}, Bochner-feature TGAT~\cite{xu2020tgat}---inform the time embedding of Section~\ref{sec:methodology}, which follows the parameterization of~\cite{linearTimeEmb2024} (originally proposed for irregular time-series forecasting). Most directly adjacent to our work, Abacus~\cite{castro2026abacus} applies self-supervised sequential user modeling to display advertising, combining a counting-aligned pretext task with masked-sequence modeling and Barlow Twins on a GRU backbone; we differ in contrastive design (NT-Xent over temporally-disjoint partitions), integration topology, and the LLM-in-the-loop optimization workflow.

\paragraph{Search over training-pipeline edits.} Random search remains a strong baseline for hyperparameter and architecture search~\cite{bergstra2012random,liu2019random}. The framing of search-atoms as reviewable code edits with literature-anchored rationales is informed by AlphaEvolve~\cite{alphaevolve2025} and the broader LLM-as-optimizer line~\cite{yang2023opro,romera2024funsearch,ma2024eureka,fernando2023promptbreeder,guo2024evoprompt,meyerson2024lmx}; AlphaEvolve and FunSearch propagate evaluation feedback into the LLM proposer, OPRO does so via a textual prior over previously-evaluated configurations, and Eureka, Promptbreeder, EvoPrompt and LMX instantiate the same closed-loop pattern over reward functions, prompts, and program candidates respectively. Section~\ref{sec:llm-optimizer} describes a two-phase instantiation: a classical hill-climbing search over the curated catalog, followed by an LLM-in-the-loop extension that receives the catalog and the running set of trial outcomes. What we borrow from this line is the closed-loop proposal structure; what we add is an industrial instantiation over a production UFM training pipeline, where the search atoms are a curated catalog of approximately $150$ literature-anchored, code-level training-pipeline edits (\emph{lifters}) whose committed sequence is inspectable and replayable.

\section{Methodology}
\label{sec:methodology}
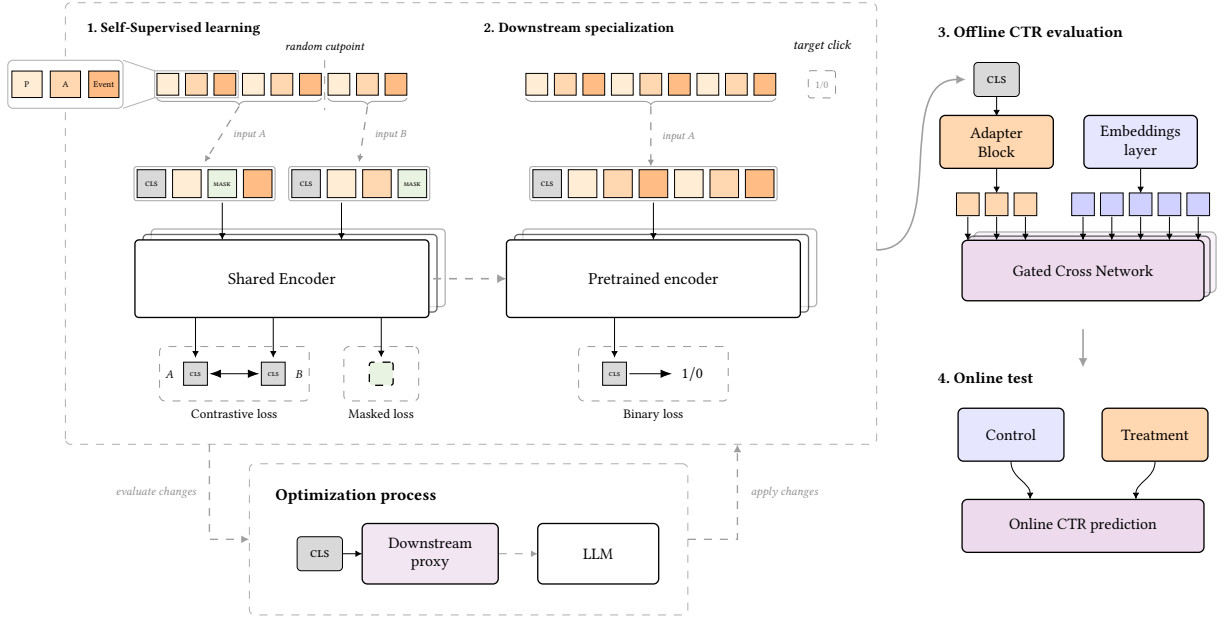
\begin{figure*}[!ht]
\centering
\begin{tikzpicture}[
  >=Latex,
  every node/.style={font=\fontsize{5}{6}\selectfont},
  smallflow/.style={->, very thin, >=Latex, shorten >=0.2pt, shorten <=0.2pt},
  rowlabel/.style={font=\fontsize{6}{7}\selectfont\bfseries, anchor=east},
  grouplabel/.style={font=\fontsize{5}{6}\selectfont, anchor=south},
  panellabel/.style={font=\scriptsize\bfseries, anchor=west},
  panellabelF/.style={font=\fontsize{6.3pt}{7.5pt}\selectfont\bfseries, anchor=west},
  tok/.style={draw, line width=0.2pt, minimum width=4.7mm, minimum height=4.6mm, inner sep=0pt, font=\fontsize{4}{5}\selectfont, align=center},
  clsT/.style={tok, fill=gray!30},
  clsTin/.style={tok, fill=gray!30, minimum width=4.2mm, minimum height=4.1mm, font=\fontsize{4}{4.8}\selectfont},
  clsTtiny/.style={tok, fill=gray!30, minimum width=3.5mm, minimum height=3.4mm, font=\fontsize{3.5}{4.2}\selectfont},
  siteT/.style={tok, fill=peachLight, minimum width=4.2mm, minimum height=4.1mm},
  advT/.style={tok, fill=peachMid,   minimum width=4.2mm, minimum height=4.1mm},
  eventT/.style={tok, fill=peachDark, minimum width=4.2mm, minimum height=4.1mm},
  maskT/.style={tok, fill=maskGreen!35, minimum width=4.2mm, minimum height=4.1mm, font=\fontsize{3.3}{4}\selectfont},
  enc/.style={draw, rounded corners=1.5pt, fill=white, line width=0.5pt, align=center},
  groupbox/.style={draw, line width=0.3pt, rounded corners=1pt, gray!70, inner sep=0pt},
  losslabel/.style={font=\fontsize{5.5}{6.5}\selectfont, anchor=north},
  lossbox/.style={draw, dashed, rounded corners=2pt, draw=gray!75, line width=0.4pt, inner sep=2pt},
  tokSmall/.style={draw, line width=0.2pt, minimum width=3.2mm, minimum height=3.2mm, inner sep=0pt},
  obox/.style={draw, rounded corners=1.5pt, fill=gray!7, align=center, font=\fontsize{6}{7}\selectfont, inner sep=1.8pt},
  box/.style={draw, rounded corners=2pt, line width=0.5pt, align=center, inner sep=2.5pt, font=\fontsize{6}{7}\selectfont},
  embox/.style={box, fill=blue!10, minimum width=15mm, minimum height=7mm},
  adabox/.style={box, fill=adapterPeach, minimum width=15mm, minimum height=7mm},
  crossbox/.style={box, fill=violet!14, minimum width=42mm, minimum height=11.4mm},
  cell/.style={draw, line width=0.2pt, minimum width=3mm, minimum height=3mm, inner sep=0pt, font=\fontsize{4}{5}\selectfont},
  cellblue/.style={cell, fill=blue!18},       
  cellorange/.style={cell, fill=adapterPeach}, 
  arr/.style={-{Latex[length=2.2pt,width=2.2pt]}, very thin},
  bigarr/.style={-{Latex[length=3.5pt,width=3.5pt]}, line width=0.6pt, gray!75},
  feedback/.style={-{Latex[length=3.0pt,width=3.0pt]}, dashed, line width=0.5pt, gray!75},
  ctrlbox/.style={box, fill=blue!10, minimum width=14mm, minimum height=7mm},
  treatbox/.style={box, fill=adapterPeach, minimum width=14mm, minimum height=7mm},
  predbox/.style={box, fill=violet!14, minimum width=32mm, minimum height=7mm},
  optblock/.style={box, fill=violet!10, minimum width=18mm, minimum height=8mm},
  x=1.27cm, y=1.27cm,
]

\draw[draw=gray!60, dashed, rounded corners=2pt, line width=0.4pt]
  (-0.50, -3.30) rectangle (7.95, 1.30);

\begin{scope}[scale=0.9, transform shape]

\node[panellabel] at (-0.4, 1.15) {1. Self-Supervised learning};

\node[font=\fontsize{5}{6}\selectfont\itshape, anchor=south] at (2.445, 0.78) {random cutpoint};
\node[tokSmall, fill=peachLight] (Ls0) at (0.63, 0.50) {};
\node[tokSmall, fill=peachMid]   (Ls1) at (0.96, 0.50) {};
\node[tokSmall, fill=peachDark]   (Ls2) at (1.29, 0.50) {};
\node[tokSmall, fill=peachLight] (Ls3) at (1.62, 0.50) {};
\node[tokSmall, fill=peachMid]   (Ls4) at (1.95, 0.50) {};
\node[tokSmall, fill=peachDark]   (Ls5) at (2.28, 0.50) {};
\node[tokSmall, fill=peachLight] (Ls6) at (2.61, 0.50) {};
\node[tokSmall, fill=peachMid]   (Ls7) at (2.94, 0.50) {};
\node[tokSmall, fill=peachDark]   (Ls8) at (3.27, 0.50) {};
\draw[dashed, gray!75, line width=0.4pt] (2.445, 0.78) -- (2.445, 0.20);

\node[clsTin] (LTi0)  at (0.44, -0.65) {\textsc{cls}};
\node[siteT]  (LTi4)  at (0.85, -0.65) {};
\node[maskT]  (LTi5)  at (1.26, -0.65) {\textsc{mask}};
\node[eventT] (LTi6)  at (1.67, -0.65) {};
\node[clsTin] (LTi7)  at (2.23, -0.65) {\textsc{cls}};
\node[siteT]  (LTi8)  at (2.64, -0.65) {};
\node[advT]   (LTi9)  at (3.05, -0.65) {};
\node[maskT]  (LTi10) at (3.46, -0.65) {\textsc{mask}};

\draw[groupbox] ($(LTi0.north west)+(-1.2pt,1.2pt)$) rectangle ($(LTi6.south east)+(1.2pt,-1.2pt)$);
\draw[groupbox] ($(LTi7.north west)+(-1.2pt,1.2pt)$) rectangle ($(LTi10.south east)+(1.2pt,-1.2pt)$);

\node[enc, minimum width=43.2mm, minimum height=11.4mm, opacity=0.30] at ($(1.95,-1.75)+(2.4mm,1.6mm)$) {};
\node[enc, minimum width=43.2mm, minimum height=11.4mm, opacity=0.60] at ($(1.95,-1.75)+(1.2mm,0.8mm)$) {};
\node[enc, minimum width=43.2mm, minimum height=11.4mm] (LENC) at (1.95, -1.75) {};
\node[font=\footnotesize] at (LENC.center) {Shared Encoder};

\coordinate (LgAi) at (1.26, -0.65);
\coordinate (LgBi) at (2.64, -0.65);
\draw[arr] ($(LgAi |- LTi0.south)+(0,-2pt)$) -- (LgAi |- LENC.north);
\draw[arr] ($(LgBi |- LTi0.south)+(0,-2pt)$) -- (LgBi |- LENC.north);

\node[lossbox, minimum width=22mm, minimum height=8mm] (LCbox) at (1.40, -2.85) {};
\node[clsTtiny] (LAcls) at (0.95, -2.85) {\textsc{cls}};
\node[font=\fontsize{5}{6}\selectfont\itshape, anchor=east] at ($(LAcls.west)+(-1pt,0)$) {A};
\node[clsTtiny] (LBcls) at (1.85, -2.85) {\textsc{cls}};
\node[font=\fontsize{5}{6}\selectfont\itshape, anchor=west] at ($(LBcls.east)+(1pt,0)$) {B};
\draw[<->, very thin, >=Latex, shorten >=1pt, shorten <=1pt] (LAcls.east) -- (LBcls.west);
\node[losslabel] at (LCbox.south) {Contrastive loss};

\node[lossbox, minimum width=11mm, minimum height=8mm] (LMbox) at (3.10, -2.85) {};
\node[draw, dashed, line width=0.5pt, fill=maskGreen!35, rounded corners=1pt,
      minimum width=3.5mm, minimum height=3.4mm, inner sep=0pt] (LRmask) at (3.10, -2.85) {};
\node[losslabel] at (LMbox.south) {Masked loss};

\draw[arr] (LAcls.north |- LENC.south) -- ($(LAcls.north)+(0,2pt)$);
\draw[arr] (LBcls.north |- LENC.south) -- ($(LBcls.north)+(0,2pt)$);
\draw[arr] (LRmask.north |- LENC.south) -- ($(LRmask.north)+(0,2pt)$);

\draw[decorate, decoration={brace, amplitude=2.5pt, raise=1pt, mirror}, gray!75, line width=0.4pt]
  (Ls0.south west) -- (Ls5.south east);
\draw[decorate, decoration={brace, amplitude=2.5pt, raise=1pt, mirror}, gray!75, line width=0.4pt]
  (Ls6.south west) -- (Ls8.south east);
\draw[-{Latex[length=2.5pt,width=2.5pt]}, dashed, gray!80, line width=0.5pt]
  (1.455, 0.245) --
  node[font=\fontsize{4.5}{5.5}\selectfont\itshape, midway, right=1.5pt, gray!95] {input A}
  ($(LTi0.north)!0.5!(LTi6.north)+(0,3pt)$);
\draw[-{Latex[length=2.5pt,width=2.5pt]}, dashed, gray!80, line width=0.5pt]
  (2.94, 0.245) --
  node[font=\fontsize{4.5}{5.5}\selectfont\itshape, midway, right=1.5pt, gray!95] {input B}
  ($(LTi7.north)!0.5!(LTi10.north)+(0,3pt)$);

\begin{scope}[shift={(4.6,0)}]
\node[panellabel] at (-0.4, 1.15) {2. Downstream specialization};

\node[tokSmall, fill=peachLight] (Ms0) at (0.30, 0.50) {};
\node[tokSmall, fill=peachMid]   (Ms1) at (0.63, 0.50) {};
\node[tokSmall, fill=peachDark]   (Ms2) at (0.96, 0.50) {};
\node[tokSmall, fill=peachLight] (Ms3) at (1.29, 0.50) {};
\node[tokSmall, fill=peachMid]   (Ms4) at (1.62, 0.50) {};
\node[tokSmall, fill=peachDark]   (Ms5) at (1.95, 0.50) {};
\node[tokSmall, fill=peachLight] (Ms6) at (2.28, 0.50) {};
\node[tokSmall, fill=peachMid]   (Ms7) at (2.61, 0.50) {};
\node[tokSmall, fill=peachDark]   (Ms8) at (2.94, 0.50) {};
\node[font=\fontsize{5.5}{6.5}\selectfont\itshape, anchor=south] at (3.60, 0.78) {target click};
\node[draw, dashed, rounded corners=1pt, gray!85, line width=0.35pt,
      minimum width=4mm, minimum height=4mm, inner sep=0pt,
      font=\fontsize{4.5}{5.5}\selectfont] (Mtarget) at (3.60, 0.50) {1/0};

\draw[decorate, decoration={brace, amplitude=2.5pt, raise=1pt, mirror}, gray!75, line width=0.4pt]
  (Ms0.south west) -- (Ms8.south east);
\draw[-{Latex[length=2.5pt,width=2.5pt]}, dashed, gray!80, line width=0.5pt]
  (1.62, 0.245) --
  node[font=\fontsize{4.5}{5.5}\selectfont\itshape, midway, right=1.5pt, gray!95] {input A}
  (1.62, -0.45);

\node[clsTin] (MTi0) at (0.42, -0.65) {\textsc{cls}};
\node[siteT]  (MTi1) at (0.83, -0.65) {};
\node[advT]   (MTi2) at (1.24, -0.65) {};
\node[eventT] (MTi3) at (1.65, -0.65) {};
\node[siteT]  (MTi4) at (2.06, -0.65) {};
\node[advT]   (MTi5) at (2.47, -0.65) {};
\node[eventT] (MTi6) at (2.88, -0.65) {};

\draw[groupbox] ($(MTi0.north west)+(-1.2pt,1.2pt)$) rectangle ($(MTi6.south east)+(1.2pt,-1.2pt)$);

\node[enc, minimum width=43.2mm, minimum height=11.4mm, opacity=0.30] at ($(1.65,-1.75)+(2.4mm,1.6mm)$) {};
\node[enc, minimum width=43.2mm, minimum height=11.4mm, opacity=0.60] at ($(1.65,-1.75)+(1.2mm,0.8mm)$) {};
\node[enc, minimum width=43.2mm, minimum height=11.4mm] (MENC) at (1.65, -1.75) {};
\node[font=\footnotesize] at (MENC.center) {Pretrained encoder};

\coordinate (MgAi) at ($(MTi0)!0.5!(MTi6)$);
\draw[arr] ($(MgAi |- MTi0.south)+(0,-2pt)$) -- (MgAi |- MENC.north);

\node[lossbox, minimum width=22mm, minimum height=8mm] (Mbinbox) at (1.65, -2.85) {};
\node[clsTtiny] (Mcls) at (1.20, -2.85) {\textsc{cls}};
\node[font=\fontsize{6}{7}\selectfont] (Mone) at (2.10, -2.85) {$1/0$};
\draw[->, very thin, >=Latex, shorten >=1pt, shorten <=1pt] (Mcls.east) -- (Mone.west);
\node[losslabel] at (Mbinbox.south) {Binary loss};

\draw[arr] (Mcls.north |- MENC.south) -- ($(Mcls.north)+(0,2pt)$);
\end{scope}

\draw[bigarr, dashed] (3.70, -1.75) -- (4.55, -1.75);

\end{scope}  

\draw[gray!60, line width=0.3pt] (0.43, 0.59) rectangle (1.30, 0.31);
\draw[draw=gray!60, fill=white, rounded corners=2pt, line width=0.4pt]
  (-1.10, 0.20) rectangle (0.10, 0.70);
\node[draw, line width=0.2pt, fill=peachLight,
      minimum width=4mm, minimum height=3.5mm, inner sep=0pt,
      font=\fontsize{3.3}{4.3}\selectfont, align=center] at (-0.90, 0.45) {P};
\node[draw, line width=0.2pt, fill=peachMid,
      minimum width=4mm, minimum height=3.5mm, inner sep=0pt,
      font=\fontsize{3.3}{4.3}\selectfont, align=center] at (-0.50, 0.45) {A};
\node[draw, line width=0.2pt, fill=peachDark,
      minimum width=4mm, minimum height=3.5mm, inner sep=0pt,
      font=\fontsize{3.3}{4.3}\selectfont, align=center] at (-0.10, 0.45) {Event};
\draw[gray!60, line width=0.3pt] (0.10, 0.70) -- (0.43, 0.59);
\draw[gray!60, line width=0.3pt] (0.10, 0.20) -- (0.43, 0.31);

\begin{scope}[shift={(8.9,1.25)}]
\node[panellabel] at (-0.4, -0.25) {3. Offline CTR evaluation};

\node[draw, rounded corners=1.5pt, fill=gray!30, line width=0.25pt,
      minimum width=6mm, minimum height=4.5mm, inner sep=0pt,
      font=\fontsize{5.5}{6.5}\selectfont] (Hclsbox) at (0.30, -0.75) {\textsc{cls}};
\draw[arr] (Hclsbox.south) -- (0.30, -1.13);
\node[adabox] (Hada) at (0.30, -1.40) {Adapter\\Block};

\node[embox] (Hemb) at (1.80, -1.40) {Embeddings\\layer};

\node[cellorange] at (0.00, -2.05) {};
\node[cellorange] at (0.30, -2.05) {};
\node[cellorange] at (0.60, -2.05) {};
\node[cellblue]   at (1.20, -2.05) {};
\node[cellblue]   at (1.50, -2.05) {};
\node[cellblue]   at (1.80, -2.05) {};
\node[cellblue]   at (2.10, -2.05) {};
\node[cellblue]   at (2.40, -2.05) {};

\draw[arr] (0.30, -1.68) -- (0.30, -1.93);
\draw[arr] (1.80, -1.68) -- (1.80, -1.93);

\node[crossbox, minimum width=32mm, minimum height=8mm, opacity=0.30] at ($(1.20,-2.74)+(1.6mm,1.1mm)$) {};
\node[crossbox, minimum width=32mm, minimum height=8mm, opacity=0.60] at ($(1.20,-2.74)+(0.8mm,0.55mm)$) {};
\node[crossbox, minimum width=32mm, minimum height=8mm] (Hcross) at (1.20, -2.74) {Gated Cross Network};

\draw[arr] (0.00, -2.18) -- (0.00, -2.42);
\draw[arr] (0.30, -2.18) -- (0.30, -2.42);
\draw[arr] (0.60, -2.18) -- (0.60, -2.42);
\draw[arr] (1.20, -2.18) -- (1.20, -2.42);
\draw[arr] (1.50, -2.18) -- (1.50, -2.42);
\draw[arr] (1.80, -2.18) -- (1.80, -2.42);
\draw[arr] (2.10, -2.18) -- (2.10, -2.42);
\draw[arr] (2.40, -2.18) -- (2.40, -2.42);
\end{scope}

\begin{scope}[shift={(8.9,-2.75)}]
\node[panellabel] at (-0.4, 0.15) {4. Online test};

\node[ctrlbox]  (Octrl)  at (0.45, -0.45) {Control};
\node[treatbox] (Otreat) at (1.95, -0.45) {Treatment};
\node[predbox]  (Opred)  at (1.20, -1.40) {Online CTR prediction};

\draw[arr] (Octrl.south)  to[out=-90, in=90] ([xshift=-7mm]Opred.north);
\draw[arr] (Otreat.south) to[out=-90, in=90] ([xshift=7mm]Opred.north);
\end{scope}

\begin{scope}[shift={(1.6,-3.74)}]
\node[draw=gray!60, dashed, rounded corners=2pt, line width=0.4pt,
      minimum width=58mm, minimum height=20mm, inner sep=4pt] (Ooptbox) at (2.10, -0.55) {};
\node[panellabelF] at (0.0, -0.10) {Optimization process};
\node[draw, rounded corners=1.5pt, fill=gray!30, line width=0.25pt,
      minimum width=6mm, minimum height=4.5mm, inner sep=0pt,
      font=\fontsize{5.5}{6.5}\selectfont] (Oclsbox) at (0.55, -0.70) {\textsc{cls}};
\node[optblock] (Oproxy) at (1.70, -0.70) {Downstream\\proxy};
\node[optblock, fill=white, minimum width=16mm] (Ollm) at (3.45, -0.70) {LLM};
\draw[arr] (Oclsbox.east) -- (Oproxy.west);
\draw[arr, dashed, gray!75] (Oproxy.east) -- (Ollm.west);
\end{scope}

\draw[bigarr] (7.95, -1.275) to[out=0, in=180] ([xshift=-3pt]Hclsbox.west);
\draw[bigarr] (10.10, -2.10) -- (10.10, -2.50);

\draw[feedback] (1.0, -3.30) |- (1.415, -4.29);
\node[font=\fontsize{4.5}{5.5}\selectfont\itshape, gray!85, anchor=east]
  at (0.95, -3.80) {evaluate changes};
\draw[feedback] (5.985, -4.29) -| (6.5, -3.30);
\node[font=\fontsize{4.5}{5.5}\selectfont\itshape, gray!85, anchor=west]
  at (6.55, -3.80) {apply changes};

\end{tikzpicture}
\caption{End-to-end pipeline. (1) Self-supervised pre-training: in the input token sequence, each triplet consists of \textbf{P}~(publisher feature), \textbf{A}~(advertiser feature), and Event~(interaction type). (2) Supervised fine-tuning: the same encoder is fine-tuned on impression-level click labels. The dashed loop spans the LLM-as-optimizer search. (3) Offline CTR evaluation. (4) Online test: a 50/50 user-level A/B in production.}
\label{fig:pipeline}
\Description[Four-panel diagram of the pipeline from self-supervised pre-training to the online test, with an optimization loop.]{A user token sequence is split at a random cutpoint into two inputs, A and B, both processed by a shared Transformer encoder; the two classification (CLS) token embeddings feed a contrastive loss, and masked positions feed a masked loss. In the second panel, the pre-trained encoder receives the full sequence and is trained with a binary loss on the CLS embedding against the click target. In the third panel, the CLS embedding passes through an adapter block, is concatenated with categorical feature embeddings, and enters the Gated Cross Network. In the fourth panel, control and treatment groups feed online CTR prediction. A dashed loop labeled optimization process connects a downstream proxy and an LLM, which evaluate and apply changes to the pre-training pipeline.}
\end{figure*}

We organize the methodology around the stages through which the encoder is trained and put to use. Figure~\ref{fig:pipeline} gives the end-to-end picture; the rest of this section walks through each stage in turn. Section~\ref{sec:data} fixes notation for the user-action sequences and the views used at training time; Section~\ref{sec:pretraining} introduces the encoder and the joint masked-token and contrastive pre-training objective; Section~\ref{sec:finetuning} describes the fine-tuning stage that specializes the encoder to its first downstream target, click prediction; Section~\ref{sec:downstream} describes how the resulting pooled embedding is integrated into that downstream task. The hyperparameter choices reported below refer to the deployed configuration, which is the output of the optimization workflow described in Section~\ref{sec:llm-optimizer}; the baseline encoder is given there.

\subsection{Data construction}
\label{sec:data}
Let $S_u = [e_1, e_2, \ldots, e_{N_u}]$ denote the action sequence of user $u$, in chronological order. Each event $e_i$ is a triplet $(\mathrm{pub}_i, \mathrm{adv}_i, \mathrm{evt}_i)$ recording the publisher feature, the advertiser feature, and the user interaction type (impression, click, conversion) at timestamp $t_i$. The length $N_u$ is variable and heavy-tailed across users.

\subsubsection{Two-view construction}
The contrastive objective requires two views of the same user sequence to serve as a positive pair. Our approach uses two temporally-disjoint views of the same user history. As shown in Figure~\ref{fig:pipeline}, $S_u$ is partitioned at a random cutpoint $\kappa_u \in \{1, \ldots, N_u - 1\}$ into a past subsequence $A_u = [e_1, \ldots, e_{\kappa_u}]$ and a future subsequence $B_u = [e_{\kappa_u+1}, \ldots, e_{N_u}]$, both preserving temporal order. The shared encoder $E_\theta$ processes both views, yielding pooled $[\mathrm{CLS}]$ embeddings $c_A$ and $c_B$; the contrastive objective aligns them across users while the MLM head reconstructs masked tokens within each view independently.

A useful byproduct of this partition strategy: the pre-training corpus requires $N_u \geq 2$, so $\kappa_u = 1$ is always a valid cutpoint, creating pre-training examples in which the past subsequence $A_u$ has length one---a single event triplet. These examples expose the encoder to the history-absent regime during pre-training.
\subsubsection{Single-view construction}
As shown in Figure~\ref{fig:pipeline}, the fine-tuning stage uses only input $A$: the entire user history up to the event of interest is passed through $A$, and $B$ is discarded. The pre-trained weights transfer in full. A fine-tuning instance pairs the user's history snapshot $A_u^{(t)}$---the token sequence of all events strictly before impression timestamp $t$---with the binary click outcome $y \in \{0,1\}$ for that impression. The maximum sequence length is $L = 49$ (16 events $\times$ 3 tokens per event plus the $[\mathrm{CLS}]$ token); both $A$ and $B$ are truncated to the same $L$.

\subsubsection{Tokenization}
A fixed vocabulary $\mathcal{V}$ is built once over the distinct publisher, advertiser and interaction-type identifiers observed in the training corpus, plus three reserved tokens: $\texttt{[PAD]}$ for padding, $\texttt{[CLS]}$ prepended to every subsequence, and $\texttt{[OOV]}$ for tokens unseen during training. Each event $e_i$ contributes three consecutive token positions (publisher, advertiser, interaction type), so a user subsequence of length $N$ produces $1 + 3N$ tokens before truncation to a fixed length $L$.

\subsubsection{Time embedding}
To make the encoder sensitive to recency, each event is augmented with a continuous feature encoding its delay relative to the most recent event of the same subsequence.
Concretely, within each subsequence, every event $e_i$ carries $\Delta t_i \geq 0$ equal to the elapsed time between $e_i$ and the most recent event of that subsequence; the most recent event therefore carries $\Delta t = 0$, as does the prepended $\texttt{[CLS]}$ token.

We adopt the \emph{LinearTimeEmbedding} parameterization of~\cite{linearTimeEmb2024}, which log-scales the delay and projects it through a per-dimension learnable affine map:
\begin{equation}
\label{eq:linear-time}
e^{\text{time}}_{i,d} = \alpha_d \cdot \log\bigl(\Delta t_i / 60 + 1\bigr) + \beta_d,
\end{equation}
where $\alpha_d, \beta_d$ are scalar learnable parameters per output dimension and the division by~60 converts seconds to minutes.
Each token's input representation is the sum of three components: its token embedding, its positional embedding, and the time embedding of its parent event (shared across the three tokens of the same triplet).

\subsection{Pre-training}
\label{sec:pretraining}

\subsubsection{Encoder}
$E_\theta$ is a bidirectional Transformer encoder following BERT~\cite{vaswani2017attention,devlin2019bert}. As shown in Figure~\ref{fig:pipeline} (stage 1), it processes a tokenized subsequence and returns a pooled representation $c$ at the $[\mathrm{CLS}]$ position; applied to inputs $A$ and $B$, we write $c_A$ and $c_B$. LayerNorm is applied before each self-attention and feed-forward sub-layer; the encoder uses GELU activations and no dropout. 

\subsubsection{Masked language modeling}
Following BERT~\cite{devlin2019bert}, a fraction $\rho$ of token positions is sampled uniformly at random across the three slot families (publisher, advertiser, event-type), replaced by $[\mathrm{MASK}]$, and the encoder is trained to recover them via cross-entropy over the vocabulary. The loss is computed independently for both $A$ and $B$. We use $\rho = 0.30$, empirically chosen and higher than the BERT default of $0.15$~\cite{devlin2019bert}.

\subsubsection{Sequence-level contrastive objective}
For a mini-batch of $N$ users with pooled embeddings $\{(c_A^i, c_B^i)\}_{i=1}^{N}$, the pair $(c_A^i, c_B^i)$ is positive; all $2(N-1)$ embeddings from other users act as negatives. We minimize the normalized temperature-scaled cross-entropy loss~\cite{chen2020simclr,oord2018cpc}:
\begin{equation}
\label{eq:ntxent-half}
\ell_i^{A \to B} \;=\;
-\log
\frac{
  \exp\bigl(\mathrm{sim}(c_A^i,\, c_B^i) / \tau\bigr)
}{
  \displaystyle\sum_{j \neq i}\bigl[
    \exp\bigl(\mathrm{sim}(c_A^i,\, c_A^j) / \tau\bigr)
    + \exp\bigl(\mathrm{sim}(c_A^i,\, c_B^j) / \tau\bigr)
  \bigr]
},
\end{equation}
with the symmetric term $\ell_i^{B \to A}$ defined by swapping $c_A^i \leftrightarrow c_B^i$. The full loss averages both directions over the batch:
\begin{equation}
\label{eq:ntxent}
\mathcal{L}_{\text{NT-Xent}}
\;=\;
\frac{1}{2N}\sum_{i=1}^{N} \bigl(\ell_i^{A \to B} + \ell_i^{B \to A}\bigr).
\end{equation}
Here $\mathrm{sim}(a,b) = a^{\top} b / (\lVert a\rVert \lVert b\rVert)$ is cosine similarity, $\tau = 0.04$, and $N = 1024$ histories per mini-batch.

\subsubsection{Total objective}
The encoder is trained end-to-end against the unweighted sum
\begin{equation}
\label{eq:pretrain-total}
\mathcal{L}_{\text{pretrain}}
=
\mathcal{L}_{\text{MLM}}^A + \mathcal{L}_{\text{MLM}}^B + \mathcal{L}_{\text{NT-Xent}}.
\end{equation}
The two objectives constrain different parts of the representation: MLM forces the per-position hidden states to recover masked tokens from local context, while the contrastive term forces the pooled $[\mathrm{CLS}]$ summary of each subsequence to align with its partner across users.

\subsection{Fine-tuning}
\label{sec:finetuning}
Fine-tuning operates on an impression-level CTR dataset indexed by the user $u$, the impression timestamp $t$, and a binary click outcome $y \in \{0, 1\}$, paired with the user's history snapshot at time $t$, i.e.\ the token sequence $A_u^{(t)}$ that the encoder would have seen had this impression been the cut-point at training time.

The encoder $\theta$ is initialized from the pre-trained weights, and a single dense $+$ sigmoid click head $(w_{\text{ctr}}, b_{\text{ctr}})$ is attached on top of the pooled embedding:
\begin{equation}
\label{eq:finetune-pred}
\hat{p}_{u, t} \;=\; \sigma\bigl(w_{\text{ctr}}^{\top} \, E_\theta(A_u^{(t)})_0 \;+\; b_{\text{ctr}}\bigr).
\end{equation}
Only subsequence $A$ is consumed at this stage. The whole stack (encoder $+$ click head) is fine-tuned end-to-end against a binary cross-entropy objective. After fine-tuning, the deployed artifact is the encoder $\theta$ alone: at serving time the pooled representation $c = E_\theta(A_u^{(t)})_0$ is exported as the user sequence feature consumed by the downstream CTR ranker (Section~\ref{sec:downstream}). The click head $(w_{\text{ctr}}, b_{\text{ctr}})$ is discarded at inference.

\subsection{Embedding integration}
\label{sec:downstream}
\label{sec:adapter}
The pooled $[\mathrm{CLS}]$ embedding $c$ is fed into a feed-forward adapter block $A: \mathbb{R}^{d_{\text{model}}} \to \mathbb{R}^{d_{\text{out}}}$ that takes $c$ as its sole input. The adapter output is concatenated with the ranker's flatten layer immediately after the GDCN~\cite{wang2023gdcn} embedding layer and consumed by the unmodified gated cross network. The same protocol is applied without modification to additional downstream models evaluated in Section~\ref{sec:generalization}. We find empirically that the adapter benefits from at least two non-linear layers: a single linear projection yields no meaningful offline improvement over the unextended production CTR ranker.

\section{LLM-as-Optimizer}
\label{sec:llm-optimizer}

The architectural and training-recipe surface of a sequential foundation model is large: embedding dimension, depth and width, optimizer family, masking ratio, contrastive temperature. We improve the foundation model of Section~\ref{sec:methodology} with an LLM-in-the-loop NAS over training-pipeline edits. The unit of search is a \emph{lifter}: a reviewable code modification with a literature anchor, replacing the more usual hyperparameter sample. Lifters range from scalar recipe changes (learning rate, masking ratio) to structural edits (normalization placement, activation family, attention-head count, depth). A \emph{candidate} is a configuration produced by applying one or more lifters to the running baseline; a \emph{committed lifter} is one whose candidate improved the proxy score and was kept on the search trajectory. This unit keeps the trajectory inspectable and the resulting configuration reproducible by replaying the committed lifters in order. The framing is informed by AlphaEvolve~\cite{alphaevolve2025}, OPRO~\cite{yang2023opro}, and FunSearch~\cite{romera2024funsearch}, with related closed-loop patterns in Eureka~\cite{ma2024eureka}, Promptbreeder~\cite{fernando2023promptbreeder}, EvoPrompt~\cite{guo2024evoprompt} and LMX~\cite{meyerson2024lmx}. Claude Opus~4.7 (Anthropic) serves as the in-loop proposer for the pre-training search, which produced the final architecture. We evaluate the workflow by its outcome: a deployed encoder configuration whose gains are confirmed by downstream re-evaluation (Section~\ref{sec:offline-pretrain}) and by a live A/B test (Section~\ref{sec:live-ab}). A comparison against alternative search procedures over the same catalog is left to future work (Section~\ref{sec:limitations}).

\subsection{Search loop}
\label{sec:approach}

\subsubsection{Evaluation protocol}
\label{sec:eval-protocol}
For clarity in the rest of the paper, we fix the following dataset terminology. The \emph{pre-training corpus} is a single unlabeled dataset over user histories, used in one piece by the pre-training stage. The \emph{fine-tuning corpus} consists of two labeled splits drawn forward-in-time from the same labeled stream, the fine-tuning train set and the fine-tuning validation set. The \emph{test set} is a third forward-in-time split on which the proxy classifier introduced below is evaluated; it is disjoint from any window used to train any stage of the pipeline.

Three numbers are tracked per iteration. The primary metric is Relative Information Gain (RIG), defined as the relative log-loss reduction of the model against a constant base-rate predictor. Writing the per-impression binary log-loss of a probability assignment $\hat p$ on a labeled set $\{(y_i, \hat p_i)\}_{i=1}^{N}$ as
\begin{equation}
\label{eq:ll}
\mathrm{LL}(\hat p) \;=\; -\frac{1}{N}\sum_{i=1}^{N} \Bigl[y_i \log \hat p_i + (1 - y_i)\log(1 - \hat p_i)\Bigr],
\end{equation}
and letting $\bar y = \tfrac{1}{N}\sum_{i=1}^{N} y_i$ denote the empirical base rate on that set, RIG is
\begin{equation}
\label{eq:rig}
\mathrm{RIG}
\;=\;
\frac{\mathrm{LL}(\bar y) \;-\; \mathrm{LL}(\hat p_{\text{model}})}{\mathrm{LL}(\bar y)}
\;=\;
1 \;-\; \frac{\mathrm{LL}(\hat p_{\text{model}})}{\mathrm{LL}(\bar y)},
\end{equation}
where the base-rate term uses the constant prediction $\hat p_i \equiv \bar y$ for all $i$. ``RIG lift'' between two models is the relative improvement in RIG, in percent: the additive difference divided by the baseline RIG value. The second metric is AUC, computed on both the fine-tuning validation set and the test set. The third is training time, used by the harness as a cost proxy that rejects lifters violating a training-budget cap.

As depicted in Figure~\ref{fig:pipeline}, when a candidate's proxy test scores yield a positive RIG lift over the control model, its produced embeddings are re-evaluated against the production CTR model (using the production ranker without embedding features as the reference) on a separate offline production-aligned experimentation setup. We treat this as a confirmatory test rather than a search target.

\subsubsection{Framework}
The search pipeline is implemented in JAX with Flax and Optax. JAX is chosen for its fast just-in-time recompilation, which makes swapping training-pipeline components inside the iterative loop cheap, and for the ergonomics of replacing individual operations through pure-function transforms. The configuration produced by the search is later ported back to TensorFlow for serving.

\subsubsection{LLM interface}
On each iteration of the pre-training search, the LLM receives three artifacts: the lifter catalog, the running set of trial outcomes (lifter committed, resulting metrics, training time), and a short textual description of the configuration on which the next proposal must build. The catalog is organized by \emph{genome}, a partition of the search space into ten categories (Interaction, Processing, Temporal, Embedding, Objective, Optimization, Regularization, Inference, Data Augmentation, RecSys Logic), covering approximately $150$ axes in total, each with a literature anchor. The LLM proposes one or a small bundle of lifters; the harness either trains the resulting candidate, rejects it on a static sanity check (parameter-count or training-budget cap), or returns it for revision.

\subsubsection{End-to-end proxy pipeline}
A fully evaluated candidate passes through the complete pipeline, scored at three stages. The encoder is first pre-trained (MLM + NT-Xent) on the pre-training corpus and scored by pre-training loss; it is then fine-tuned on the impression-level CTR target and scored by validation RIG and AUC. Finally, the pooled $[\mathrm{CLS}]$ embeddings are exported, and a \emph{proxy} classifier (a two-layer feed-forward network with its own validation split) is trained from scratch on the click label using the embeddings as its only input (Figure~\ref{fig:pipeline}). The proxy is scored on the test set by AUC and RIG.

\subsection{Optimization results}
\label{sec:optim-results}

The search proceeds in two stages. The first optimizes the encoder's architectural and contrastive-objective hyperparameters by minimizing the pre-training loss on the pre-training corpus. The second optimizes the supervised-stage hyperparameters by maximizing the AUC on the fine-tuning validation set together with the AUC and RIG of the proxy classifier on the test set; this split therefore participates in this stage's selection.

\subsubsection{NAS pre-training results}
\label{sec:optim-pretrain}

\begin{figure}[!ht]
\centering
\includegraphics[width=\columnwidth]{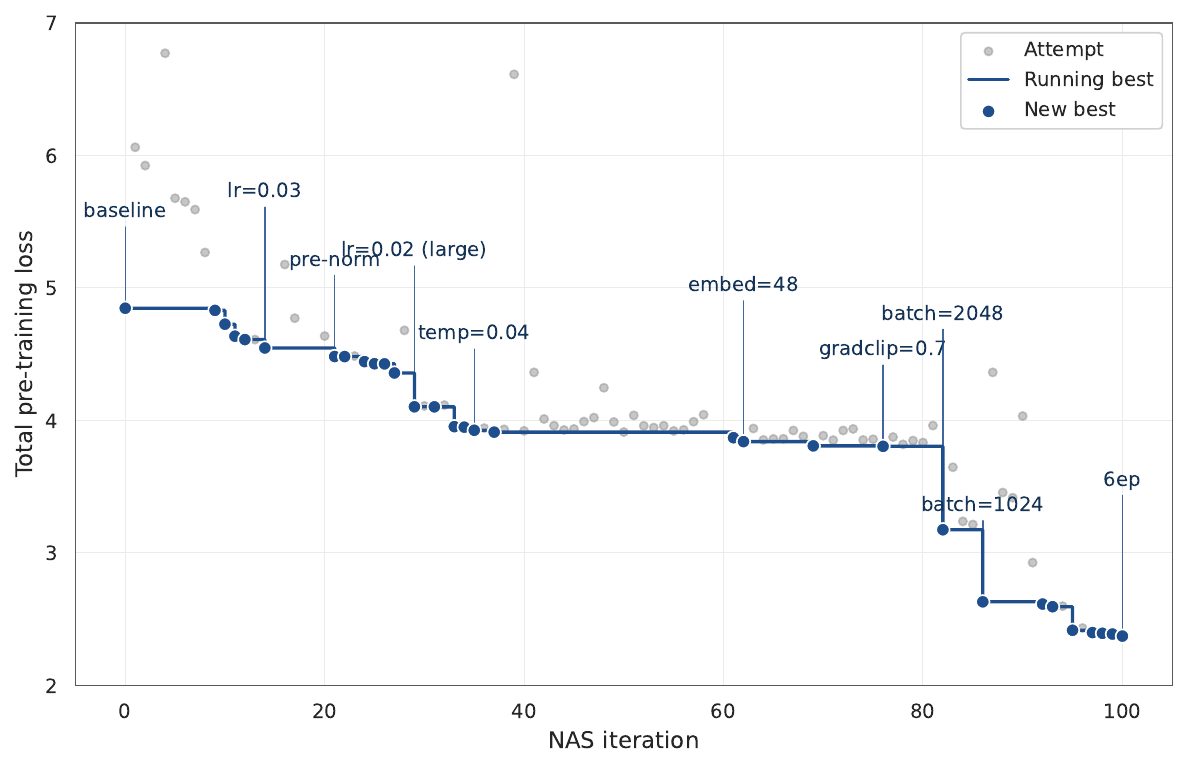}
\caption{Pre-training NAS trajectory and committed lifters.}
\label{fig:nas-pretrain-trajectory}
\Description[Total pre-training loss falls from 4.85 to 2.37 over 100 search iterations.]{Scatter plot of total pre-training loss against iteration index. Grey dots show individual candidate losses; a step curve traces the running best, decreasing from 4.85 to 2.37. Callouts mark the iterations where the best improved, including learning rate changes, the pre-norm block variant, embedding dimension 48, contrastive temperature 0.04, and batch size and epoch count changes.}
\end{figure}

Figure~\ref{fig:nas-pretrain-trajectory} traces the pre-training search across $100$ iterations (x-axis: iteration index; y-axis: total pre-training loss, MLM + NT-Xent equally weighted; grey dots: per-candidate loss; dark blue step curve: running minimum; circled markers: iterations where the minimum advanced). The running-best loss fell from $4.85$ to $2.37$ over $31$ committed candidates, in three regimes: (i)~a learning-rate sweep through iteration~$14$; (ii)~an architectural phase (iterations $21$--$76$) committing pre-norm, $8$ attention heads, $\tau{=}0.04$, embedding dimension $48$, and gradient clipping; and (iii)~a data-side phase from iteration~$82$ (batch size $1024$, $6$ training epochs). The trajectory values are proxy-run scores used for candidate selection on the search dataset; Table~\ref{tab:nas-pretrain} reports a dedicated full retrain of the committed configuration on a separate dataset.

\begin{table}[!ht]
\caption{NAS pre-training results. NAS-lifted: encoder after the NAS search.}
\label{tab:nas-pretrain}
\small
\begin{tabular}{lrrr}
\toprule
                  & Baseline ($10$ ep.) & NAS-lifted ($6$ ep.) & $\Delta$               \\
\midrule
Total loss        & $2.5566$            & $\boldsymbol{1.9113}$ & $\boldsymbol{-25.2\%}$ \\
MLM loss (per view) & $0.1847$            & $0.1348$              & $-27.0\%$              \\
Contrastive loss  & $2.1872$            & $1.6417$              & $-24.9\%$              \\
Training time     & $745.1$ min         & $764.4$ min           & $+2.6\%$               \\
Time / epoch      & $\sim$$74.5$ min    & $\sim$$127.4$ min     & ---                    \\
\bottomrule
\end{tabular}
\end{table}

Table~\ref{tab:nas-pretrain} compares the pre-training loss of the UFM before the optimization loop---the Baseline, pre-trained for $10$ epochs---and the NAS-lifted version, the encoder after optimization. NAS-lifted reduces total pre-training loss by $-25.2\%$ relative to the Baseline. Despite a larger model, wall-clock cost rises by only $+2.6\%$ because the NAS-lifted encoder converges in $6$ epochs.

\FloatBarrier
\subsubsection{NAS fine-tuning and downstream proxy results}
\label{sec:optim-finetune}
With the pre-training stage held at its committed configuration, the search then optimized the fine-tuning stage of the foundation model. The pooled representation of each candidate is passed to the proxy classifier (Section~\ref{sec:approach}) and scored on the test set against the click labels. Table~\ref{tab:nas-ft} reports the result, together with the skip-pre-training ablation introduced in Section~\ref{sec:optim-pretrain}.

\begin{table}[!ht]
\caption{NAS fine-tuning and proxy results on the test set.}
\label{tab:nas-ft}
\small
\setlength{\tabcolsep}{4pt}
\begin{tabular}{lrrr}
\toprule
Metric              & \shortstack[r]{Baseline\\(pre-NAS)} & \shortstack[r]{NAS-\\lifted}  & \shortstack[r]{NAS-lifted\\(skip pre-train)} \\
\midrule
Fine-tune Val AUC   & $0.7450$    & $\boldsymbol{0.7491}$   & $0.7396$  \\
Test AUC            & $0.7160$    & $\boldsymbol{0.7196}$   & $0.7148$  \\
Test RIG            & $0.0510$    & $\boldsymbol{0.0547}$   & $0.0467$  \\
\bottomrule
\end{tabular}
\end{table}

The NAS-lifted configuration improves test RIG by $+7.3\%$ relative over the baseline encoder on the same test set. Removing the pre-training stage on the same architecture drops test RIG to $0.0467$, a $-14.6\%$ relative regression from the NAS-lifted configuration that also falls below the baseline encoder RIG of $0.0510$, confirming that the MLM + NT-Xent stage contributes materially and that the lift is attributable to the pre-training objective.

\section{Results}
\label{sec:results}

Results are organized around four questions:
\begin{itemize}
  \item \textbf{RQ1.} Does the UFM produce a user representation that materially improves the production CTR metric (RIG) over the unextended GDCN ranker, across different history availability regimes?
  \item \textbf{RQ2.} Does the improvement generalize across ranker architectures and prediction tasks---constituting a ranker-agnostic and task-agnostic improvement axis?
  \item \textbf{RQ3.} Does the LLM-in-the-loop NAS produce a deployable encoder configuration, and what does each phase contribute?
  \item \textbf{RQ4.} Is the deployed encoder feasible under the latency budget of C1, and how does its serving topology handle opt-out requests?
\end{itemize}
RQ1 is addressed in Section~\ref{sec:offline-pretrain}; RQ2 in Section~\ref{sec:generalization}; RQ3 in Section~\ref{sec:optim-results} and the live test of Section~\ref{sec:live-ab}; RQ4 in Section~\ref{sec:serving}.

\subsection{Offline downstream results}
\label{sec:offline-pretrain}

\begin{table}[!ht]
\caption{Offline downstream CTR results in RIG against the production GDCN ranker.}
\label{tab:offline-downstream}
\small
\begin{tabular}{lr}
\toprule
Configuration                        & RIG $\Delta$ \\
\midrule
GDCN                                 & ---          \\
GDCN + UFM-Base                      & $+1.065\%$   \\
GDCN + UFM-NAS                       & $\boldsymbol{+1.354\%}$ \\
\bottomrule
\end{tabular}
\end{table}

We compare three configurations. The reference is the production GDCN without any user embedding extension. The first UFM variant is the baseline encoder integrated through the deployed adapter. The second is the NAS-lifted encoder integrated with the same protocol. Table~\ref{tab:offline-downstream} reports RIG relative to the unextended reference (UFM-Base: encoder trained before the NAS search; UFM-NAS: deployed NAS-lifted encoder): UFM-Base yields $+1.065\%$ RIG and UFM-NAS $+1.354\%$ RIG, with the optimization loop accounting for the gap between the two.

\begin{figure}[!ht]
\centering
\includegraphics[width=\columnwidth]{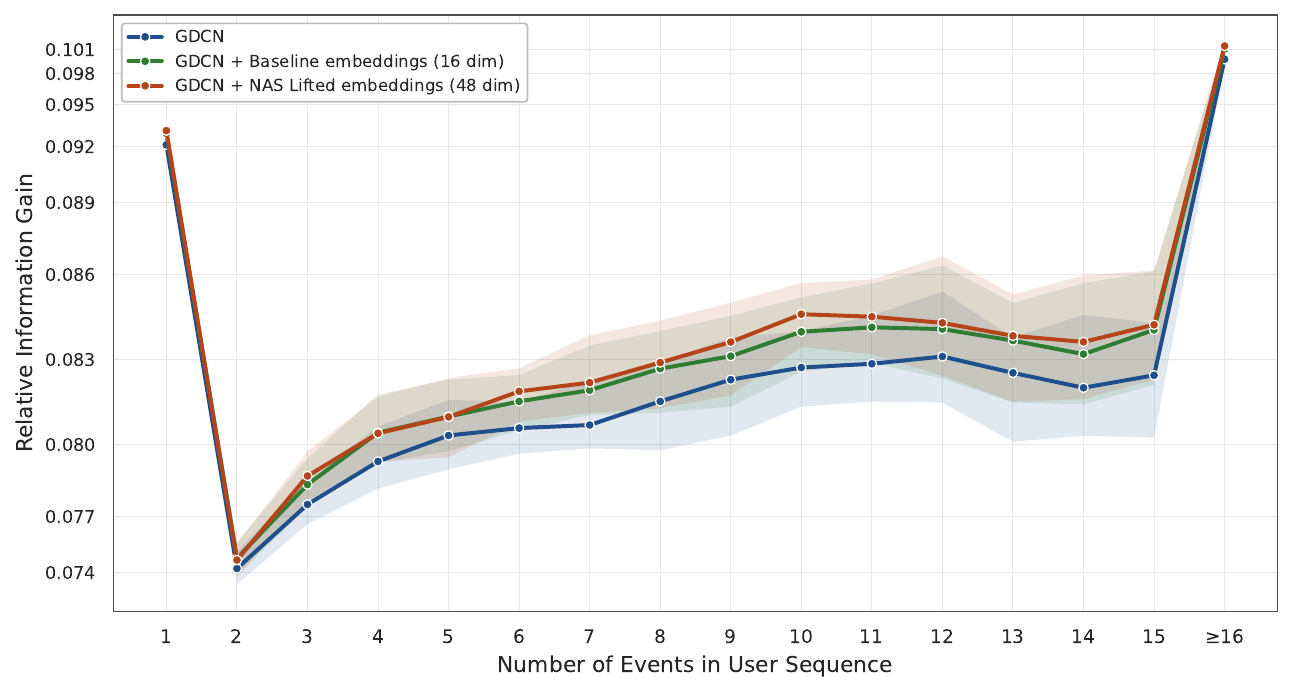}
\caption{RIG broken down by number of events in the user sequence.}
\label{fig:rig-by-length}
\Description[RIG dips at two events, rises with history length, and jumps at sixteen or more; the embedding variants stay above the GDCN reference.]{Line chart of Relative Information Gain against the number of events in the user history sequence, from one to sixteen or more, for three models: the GDCN reference, GDCN with 16-dimensional baseline embeddings, and GDCN with 48-dimensional NAS-lifted embeddings. All three curves start near 0.092 at one event, drop to a minimum of about 0.074 at two events, rise gradually to about 0.084 by fifteen events, and jump to about 0.101 at sixteen or more. The embedding variants lie above the reference throughout, with the NAS-lifted variant generally highest and the gap most visible from about six events onward; shaded bands indicate variability.}
\end{figure}

Figure~\ref{fig:rig-by-length} breaks the headline lift down by the number of events (triplets) in the user history sequence. (i)~At history length~$1$, this bucket consists solely of the current event triplet, with no prior browsing history. Here, UFM-NAS yields $+0.99\%$ relative RIG over the GDCN reference. (ii)~Starting from 2 events (2 triplets), the lift of both UFM-Base and UFM-NAS becomes more prominent; UFM-NAS pulls further ahead of UFM-Base, particularly from 6 events per sequence onward. The total RIG lift on the history-present portion of traffic reaches $+1.62\%$ RIG. Table~\ref{tab:rig-by-length} summarizes the two aggregate buckets.

\begin{table}[t]
\caption{UFM-NAS relative RIG lift over the GDCN reference, by number of events in the sequence.}
\label{tab:rig-by-length}
\small
\begin{tabular}{lr}
\toprule
History length & UFM-NAS vs.\ GDCN \\
\midrule
$= 1$ (opt-out dominated) & $+0.99\%$ \\
$\geq 2$                  & $+1.62\%$ \\
\bottomrule
\end{tabular}
\end{table}

\subsection{Ranker and task generalization}
\begin{table}[!ht]
\caption{Offline RIG lift across ranker architectures and prediction tasks.}
\label{tab:generalization}
\small
\begin{tabular}{llr}
\toprule
Downstream model  & Prediction task  & RIG $\Delta$ \\
\midrule
GDCN              & CTR              & $\boldsymbol{+1.354\%}$ \\
DCN$^2$           & CTR              & $+0.894\%$ \\
Win-rate model    & Win probability  & $+1.197\%$ \\
\bottomrule
\end{tabular}
\end{table}

\label{sec:generalization}
The same NAS-optimized encoder representation, with the same adapter integration protocol, is evaluated on two additional models. First, a second CTR ranker architecture (DCN$^2$) is tested to assess cross-architecture portability. Although the encoder was fine-tuned on CTR prediction, its embeddings were also tested in the win-rate model to probe cross-task transfer. Fine-tuning the encoder on the win-rate task itself was not tested; the encoder is used as-is, with click-task fine-tuning only. Table~\ref{tab:generalization} reports the results. The encoder improves all three models: the two CTR rankers gain $+1.354\%$ and $+0.894\%$ RIG respectively, and the win-rate model gains $+1.197\%$ RIG, suggesting that the sequential user representation carries signal beyond click prediction and is a valuable additional feature even for structurally different tasks. We see extending fine-tuning beyond the click task as a promising direction for future work.

\subsection{Online serving}
\label{sec:serving}
Because the encoder is a user foundation model rather than a request-level feature, it is deployed in a dedicated data management platform (DMP) service that handles user-modeling inference independently from the CTR ranker. User interactions are streamed in near real-time into a low-latency key-value database (capped at $16$ events, $7$-day TTL); at each bid request the DMP fetches the visitor's most recent history, builds the token sequence, and runs the inference-optimized encoder in a micro-batched pipeline (width~$8$, maximum wait~$1$~ms). The resulting pooled $[\mathrm{CLS}]$ embedding is transmitted to the demand-side platform (DSP) alongside the standard bid-request features. The DSP then feeds it through the adapter block described in Section~\ref{sec:adapter} and runs the GDCN to produce the final CTR prediction. The full DMP-to-DSP round trip fits within the sub-millisecond bid-path budget of C1.

For opt-out users whose browsing history is unavailable, the DMP builds a sequence containing only the current bid-request triplet. This topology satisfies both operating requirements of RQ4: sub-millisecond end-to-end latency on the bid path and stable inference in the history-absent regime.

\subsection{Live A/B test}
\label{sec:live-ab}
The live A/B test ran on full production traffic over 7 days with a 50/50 split between the control and treatment groups; campaign budgets were equally allocated across both groups. Users with a persistent visitor identifier were randomized by hashed identifier and consistently routed to the same group across requests; opt-out traffic, which by construction has no cross-request identity, was randomized at the bid-request level. The control group scored bid requests with the unmodified production CTR ranker; the treatment group used the same ranker augmented by the UFM-NAS embeddings through the adapter described in Section~\ref{sec:adapter}. Table~\ref{tab:ab-test} reports the results: $+2.13\%$ CTR and $-1.13\%$ eCPC on click-optimized traffic, and $+2.37\%$ visit rate (post-click site visits per served impression) and $-1.47\%$ eCPV (advertiser cost per visit) on visit-optimized traffic, all with $80\%$ confidence intervals (our internal reporting standard for production A/B tests) excluding zero, confirming that the offline RIG improvements translate to gains on production KPIs.

\begin{table}[h]
\centering
\caption{Online A/B test results over 7 days. All figures are relative differences.}
\label{tab:ab-test}
\begin{tabular}{lcc}
\toprule
Metric & Treatment $-$ Control & 80\% CI \\
\midrule
CTR & $+2.13\%$ & [$+1.55\%$, $+3.17\%$] \\
eCPC & $-1.13\%$ & [$-2.07\%$, $-0.64\%$] \\
Visit rate & $+2.37\%$ & [$+1.49\%$, $+3.31\%$] \\
eCPV & $-1.47\%$ & [$-2.59\%$, $-0.48\%$] \\
\bottomrule
\end{tabular}
\end{table}

\subsection{Current limitations}
\label{sec:limitations}
Three methodological caveats apply to the offline evaluation. (i) \textit{Single-seed point estimates.} The numbers in Tables~\ref{tab:nas-pretrain} and~\ref{tab:nas-ft} are computed from one training run per cell; we treat them as point estimates confirmed by the live A/B test rather than standalone claims. (ii) \textit{No search-procedure control.} The lift in Table~\ref{tab:nas-pretrain} measures NAS-lifted vs.\ baseline encoder, not LLM-driven search vs.\ alternative procedures over the same catalog; we cannot rule out that the curated catalog itself, rather than the LLM proposer, is the principal contribution. Isolating the two would require repeating the search under alternative proposal strategies at matched budget; we defer this controlled study to future work and accordingly scope the claim of Section~\ref{sec:llm-optimizer} as an existence proof of the workflow rather than a superiority result. (iii) \textit{Test-set vs.\ in-search evaluation.} Phase-1 NAS reported the argmax over 100 end-to-end iterations. The downstream evaluation of Section~\ref{sec:offline-pretrain} is disjoint from the selection data, but the architectural choice itself was made by maximizing over 100 noisy candidates and is therefore subject to winner's-curse bias; a multi-seed retraining of the selected configuration would tighten the estimate. The supervised-stage selection likewise consulted proxy scores on the test set; since its candidates shared one pre-trained backbone and differed only in fine-tuning recipe and evaluation-head configuration, the scope for selection-induced optimism is narrow.

\subsection{Future perspectives}
\label{sec:future}
Three directions stand out for future work. \textbf{Broader downstream tasks.} The generalization evidence (CTR, win-rate) motivates extending the same encoder to tasks with structurally different supervision signals. \textbf{Encoder architecture.} The deployed encoder is a standard bidirectional Transformer with known limitations on training time and serving latency; architectural improvements on both axes remain an open direction. \textbf{Sequence feature selection.} The current token vocabulary (publisher, advertiser, interaction type) leaves price signals, creative context, and session-level features unexplored; systematic ablation of what enters the sequence is an open question with direct impact on representation quality.

\section{Conclusion}
\label{sec:conclusion}

In this work, we demonstrate a successful deployment of a User Foundation Model at Teads under the open-web RTB conditions of strict sub-millisecond latency (C1) and non-persistent user identity with sparse, bounded history (C2). Integrating the base encoder into the production GDCN ranker yields $+1.065\%$ RIG; applying the LLM-in-the-loop NAS raises this to $+1.354\%$, demonstrating sequential modeling as a new improvement axis for downstream models at Teads. The same representation generalizes across ranker architectures ($+0.894\%$ RIG on DCN$^2$) and to a structurally different prediction task ($+1.197\%$ RIG on the bid win-rate model, with click-only fine-tuning) --- evidence that the encoder captures historical signal layered on top of the discrete, hand-crafted features the production ranker already consumes. A 7-day live A/B test against the production ranker confirms the offline picture: $+2.13\%$ CTR, $-1.13\%$ eCPC, $+2.37\%$ visit rate and $-1.47\%$ eCPV, all with 80\% confidence intervals excluding zero. Future work will extend the encoder to downstream tasks with different supervision signals and broaden the search space of the pre-training optimization.

\bibliographystyle{ACM-Reference-Format}
\bibliography{references}

@inproceedings{vaswani2017attention,
  title     = {Attention is All You Need},
  author    = {Vaswani, Ashish and Shazeer, Noam and Parmar, Niki and Uszkoreit, Jakob and Jones, Llion and Gomez, Aidan N. and Kaiser, {\L}ukasz and Polosukhin, Illia},
  booktitle = {Advances in Neural Information Processing Systems (NeurIPS)},
  year      = {2017},
  pages     = {5998--6008}
}

@inproceedings{devlin2019bert,
  title     = {{BERT}: Pre-training of Deep Bidirectional Transformers for Language Understanding},
  author    = {Devlin, Jacob and Chang, Ming-Wei and Lee, Kenton and Toutanova, Kristina},
  booktitle = {Proceedings of the 2019 Conference of the North American Chapter of the Association for Computational Linguistics (NAACL-HLT)},
  year      = {2019},
  pages     = {4171--4186}
}

@misc{linearTimeEmb2024,
  title         = {Continuous-Time Linear Positional Embedding for Irregular Time Series Forecasting},
  author        = {Kim, Byunghyun and Lee, Jae-Gil},
  year          = {2024},
  eprint        = {2409.20092},
  archivePrefix = {arXiv},
  primaryClass  = {cs.LG}
}

@inproceedings{kang2018sasrec,
  title     = {Self-Attentive Sequential Recommendation},
  author    = {Kang, Wang-Cheng and McAuley, Julian},
  booktitle = {IEEE International Conference on Data Mining (ICDM)},
  year      = {2018},
  pages     = {197--206}
}

@inproceedings{sun2019bert4rec,
  title     = {{BERT4Rec}: Sequential Recommendation with Bidirectional Encoder Representations from Transformer},
  author    = {Sun, Fei and Liu, Jun and Wu, Jian and Pei, Changhua and Lin, Xiao and Ou, Wenwu and Jiang, Peng},
  booktitle = {Proceedings of the 28th {ACM} International Conference on Information and Knowledge Management (CIKM)},
  year      = {2019},
  pages     = {1441--1450}
}

@inproceedings{xie2022cl4srec,
  title     = {Contrastive Learning for Sequential Recommendation},
  author    = {Xie, Xu and Sun, Fei and Liu, Zhaoyang and Wu, Shiwen and Gao, Jinyang and Ding, Bolin and Cui, Bin},
  booktitle = {IEEE 38th International Conference on Data Engineering (ICDE)},
  year      = {2022},
  pages     = {1259--1273}
}

@inproceedings{qiu2022duorec,
  title     = {Contrastive Learning for Representation Degeneration Problem in Sequential Recommendation},
  author    = {Qiu, Ruihong and Huang, Zi and Yin, Hongzhi and Wang, Zijian},
  booktitle = {Proceedings of the Fifteenth ACM International Conference on Web Search and Data Mining (WSDM)},
  year      = {2022},
  pages     = {813--823}
}

@inproceedings{zhou2020s3rec,
  title     = {{S$^3$-Rec}: Self-Supervised Learning for Sequential Recommendation with Mutual Information Maximization},
  author    = {Zhou, Kun and Wang, Hui and Zhao, Wayne Xin and Zhu, Yutao and Wang, Sirui and Zhang, Fuzheng and Wang, Zhongyuan and Wen, Ji-Rong},
  booktitle = {Proceedings of the 29th {ACM} International Conference on Information and Knowledge Management (CIKM)},
  year      = {2020},
  pages     = {1893--1902}
}

@article{wang2023contrarec,
  title   = {Sequential Recommendation with Multiple Contrast Signals},
  author  = {Wang, Chenyang and Ma, Weizhi and Chen, Chong and Zhang, Min and Liu, Yiqun and Ma, Shaoping},
  journal = {ACM Transactions on Information Systems (TOIS)},
  volume  = {41},
  number  = {1},
  pages   = {Article 11},
  year    = {2023}
}

@inproceedings{li2020tisasrec,
  title     = {Time Interval Aware Self-Attention for Sequential Recommendation},
  author    = {Li, Jiacheng and Wang, Yujie and McAuley, Julian},
  booktitle = {Proceedings of the Thirteenth ACM International Conference on Web Search and Data Mining (WSDM)},
  year      = {2020},
  pages     = {322--330}
}

@misc{kazemi2019time2vec,
  title         = {Time2Vec: Learning a Vector Representation of Time},
  author        = {Kazemi, Seyed Mehran and Goel, Rishab and Eghbali, Sepehr and Ramanan, Janahan and Sahota, Jaspreet and Thakur, Sanjay and Wu, Stella and Smyth, Cathal and Poupart, Pascal and Brubaker, Marcus},
  year          = {2019},
  eprint        = {1907.05321},
  archivePrefix = {arXiv},
  primaryClass  = {cs.LG}
}

@inproceedings{xu2020tgat,
  title     = {Inductive Representation Learning on Temporal Graphs},
  author    = {Xu, Da and Ruan, Chuanwei and Korpeoglu, Evren and Kumar, Sushant and Achan, Kannan},
  booktitle = {International Conference on Learning Representations (ICLR)},
  year      = {2020}
}

@inproceedings{chen2019bst,
  title     = {Behavior Sequence Transformer for {E-commerce} Recommendation in {Alibaba}},
  author    = {Chen, Qiwei and Zhao, Huan and Li, Wei and Huang, Pipei and Ou, Wenwu},
  booktitle = {Proceedings of the 1st International Workshop on Deep Learning Practice for High-Dimensional Sparse Data (DLP-KDD)},
  year      = {2019},
  pages     = {Article 12}
}

@inproceedings{zhou2018din,
  title     = {Deep Interest Network for Click-Through Rate Prediction},
  author    = {Zhou, Guorui and Zhu, Xiaoqiang and Song, Chenru and Fan, Ying and Zhu, Han and Ma, Xiao and Yan, Yanghui and Jin, Junqi and Li, Han and Gai, Kun},
  booktitle = {Proceedings of the 24th {ACM} {SIGKDD} International Conference on Knowledge Discovery \& Data Mining (KDD)},
  year      = {2018},
  pages     = {1059--1068}
}

@inproceedings{pancha2022pinnerformer,
  title     = {{PinnerFormer}: Sequence Modeling for User Representation at {Pinterest}},
  author    = {Pancha, Nikil and Zhai, Andrew and Leskovec, Jure and Rosenberg, Charles},
  booktitle = {Proceedings of the 28th {ACM} {SIGKDD} Conference on Knowledge Discovery and Data Mining (KDD)},
  year      = {2022},
  pages     = {3702--3712}
}

@misc{shin2021one4all,
  title         = {One4all User Representation for Recommender Systems in {E}-commerce},
  author        = {Shin, Kyuyong and Kwak, Hanock and Kim, Kyung-Min and Kim, Minkyu and Park, Young-Jin and Jeong, Jisu and Jung, Seungjae},
  year          = {2021},
  eprint        = {2106.00573},
  archivePrefix = {arXiv},
  primaryClass  = {cs.IR},
  note          = {ShopperBERT}
}

@inproceedings{yang2023lurm,
  title     = {Empowering General-Purpose User Representation with Full-Life-Cycle Behavior Modeling},
  author    = {Yang, Bei and Gu, Jie and Liu, Ke and Xu, Xiaoxiao and Xu, Renjun and Sun, Qinghui and Liu, Hong},
  booktitle = {Proceedings of the 29th {ACM} {SIGKDD} Conference on Knowledge Discovery and Data Mining (KDD)},
  year      = {2023},
  doi       = {10.1145/3580305.3599331},
  note      = {arXiv:2110.11337}
}

@misc{chen2024hllm,
  title         = {{HLLM}: Enhancing Sequential Recommendations via Hierarchical Large Language Models for Item and User Modeling},
  author        = {Chen, Junyi and Chi, Lu and Peng, Bingyue and Yuan, Zehuan},
  year          = {2024},
  eprint        = {2409.12740},
  archivePrefix = {arXiv},
  primaryClass  = {cs.IR}
}

@inproceedings{chen2025pinfm,
  title     = {{PinFM}: Foundation Model for User Activity Sequences at a Billion-Scale Visual Discovery Platform},
  author    = {Chen, Xiangyi and Rajesh, Kousik and Lawhon, Matthew and Wang, Zelun and Li, Hanyu and Li, Haomiao and Joshi, Saurabh Vishwas and Eksombatchai, Pong and Yang, Jaewon and Hsu, Yi-Ping and Xu, Jiajing and Rosenberg, Charles},
  booktitle = {Proceedings of the 19th {ACM} Conference on Recommender Systems (RecSys)},
  year      = {2025},
  doi       = {10.1145/3705328.3748050}
}

@inproceedings{liang2025elfm,
  title     = {External Large Foundation Model: How to Efficiently Serve Trillions of Parameters for Online Ads Recommendation},
  author    = {Liang, Mingfu and Liu, Xi and Jin, Rong and Liu, Boyang and Suo, Qiuling and Zhou, Qinghai and Zhou, Song and Chen, Laming and Zheng, Hua and Li, Zhiyuan and others},
  booktitle = {Companion Proceedings of the {ACM} on Web Conference 2025 ({WWW} Industry Track)},
  year      = {2025},
  note      = {arXiv:2502.17494}
}

@misc{song2026mtfm,
  title         = {{MTFM}: A Scalable and Alignment-Free Foundation Model for Industrial Recommendation in {Meituan}},
  author        = {Song, Xin and Guan, Zhilin and Han, Ruidong and others},
  year          = {2026},
  eprint        = {2602.11235},
  archivePrefix = {arXiv},
  primaryClass  = {cs.IR}
}

@inproceedings{gao2021simcse,
  title     = {{SimCSE}: Simple Contrastive Learning of Sentence Embeddings},
  author    = {Gao, Tianyu and Yao, Xingcheng and Chen, Danqi},
  booktitle = {Proceedings of the 2021 Conference on Empirical Methods in Natural Language Processing (EMNLP)},
  year      = {2021},
  pages     = {6894--6910}
}

@article{bergstra2012random,
  title   = {Random Search for Hyper-Parameter Optimization},
  author  = {Bergstra, James and Bengio, Yoshua},
  journal = {Journal of Machine Learning Research},
  volume  = {13},
  number  = {2},
  pages   = {281--305},
  year    = {2012}
}

@inproceedings{liu2019random,
  title     = {Random Search and Reproducibility for Neural Architecture Search},
  author    = {Li, Liam and Talwalkar, Ameet},
  booktitle = {Proceedings of the 35th Conference on Uncertainty in Artificial Intelligence (UAI)},
  year      = {2019},
  pages     = {367--377}
}

@inproceedings{romera2024funsearch,
  title     = {Mathematical Discoveries from Program Search with Large Language Models},
  author    = {Romera-Paredes, Bernardino and Barekatain, Mohammadamin and Novikov, Alexander and Balog, Matej and Kumar, M. Pawan and Dupont, Emilien and Ruiz, Francisco J. R. and Ellenberg, Jordan S. and Wang, Pengming and Fawzi, Omar and Kohli, Pushmeet and Fawzi, Alhussein},
  booktitle = {Nature},
  year      = {2024},
  pages     = {468--475},
  volume    = {625}
}

@inproceedings{chen2020simclr,
  title     = {A Simple Framework for Contrastive Learning of Visual Representations},
  author    = {Chen, Ting and Kornblith, Simon and Norouzi, Mohammad and Hinton, Geoffrey},
  booktitle = {International Conference on Machine Learning (ICML)},
  year      = {2020},
  pages     = {1597--1607}
}

@misc{oord2018cpc,
  title         = {Representation Learning with Contrastive Predictive Coding},
  author        = {van den Oord, A{\"a}ron and Li, Yazhe and Vinyals, Oriol},
  year          = {2018},
  eprint        = {1807.03748},
  archivePrefix = {arXiv},
  primaryClass  = {cs.LG}
}

@misc{wang2025dcn2,
  title         = {{DCN}$^2$: Interplay of Implicit Collision Weights and Explicit Cross Layers for Large-Scale Recommendation},
  author        = {\v{S}krlj, Bla\v{z} and Karni, Yonatan and Ga\v{s}per\v{s}i\v{c}, Grega and Mramor, Bla\v{z} and Stolin, Yulia and Jakomin, Martin and Urban\v{c}i\v{c}, Jasna and Dishi, Yuval and Silberstein, Natalia and Friedler, Ophir and Klein, Assaf},
  year          = {2025},
  eprint        = {2506.21624},
  archivePrefix = {arXiv},
  primaryClass  = {cs.LG}
}

@inproceedings{wang2023gdcn,
  title     = {Towards Deeper, Lighter and Interpretable Cross Network for {CTR} Prediction},
  author    = {Wang, Fangye and Gu, Hansu and Li, Dongsheng and Lu, Tun and Zhang, Peng and Gu, Ning},
  booktitle = {Proceedings of the 32nd {ACM} International Conference on Information and Knowledge Management (CIKM)},
  year      = {2023},
  pages     = {2628--2637},
  publisher = {ACM},
  doi       = {10.1145/3583780.3615089},
  note      = {arXiv:2311.04635}
}

@misc{alphaevolve2025,
  title         = {{AlphaEvolve}: A Coding Agent for Scientific and Algorithmic Discovery},
  author        = {Novikov, Alexander and V{\~u}, Ng{\^a}n and Eisenberger, Marvin and Dupont, Emilien and Huang, Po-Sen and Wagner, Adam Zsolt and others},
  year          = {2025},
  eprint        = {2506.13131},
  archivePrefix = {arXiv},
  primaryClass  = {cs.AI},
  note          = {Google DeepMind technical report}
}

@inproceedings{yang2023opro,
  title     = {Large Language Models as Optimizers},
  author    = {Yang, Chengrun and Wang, Xuezhi and Lu, Yifeng and Liu, Hanxiao and Le, Quoc V. and Zhou, Denny and Chen, Xinyun},
  booktitle = {International Conference on Learning Representations (ICLR)},
  year      = {2024},
  note      = {arXiv:2309.03409}
}

@inproceedings{ma2024eureka,
  title     = {Eureka: Human-Level Reward Design via Coding Large Language Models},
  author    = {Ma, Yecheng Jason and Liang, William and Wang, Guanzhi and Huang, De-An and Bastani, Osbert and Jayaraman, Dinesh and Zhu, Yuke and Fan, Linxi and Anandkumar, Anima},
  booktitle = {International Conference on Learning Representations (ICLR)},
  year      = {2024},
  note      = {arXiv:2310.12931}
}

@misc{fernando2023promptbreeder,
  title         = {Promptbreeder: Self-Referential Self-Improvement via Prompt Evolution},
  author        = {Fernando, Chrisantha and Banarse, Dylan and Michalewski, Henryk and Osindero, Simon and Rockt{\"a}schel, Tim},
  year          = {2023},
  eprint        = {2309.16797},
  archivePrefix = {arXiv}
}

@inproceedings{guo2024evoprompt,
  title     = {Connecting Large Language Models with Evolutionary Algorithms Yields Powerful Prompt Optimizers},
  author    = {Guo, Qingyan and Wang, Rui and Guo, Junliang and Li, Bei and Song, Kaitao and Tan, Xu and Liu, Guoqing and Bian, Jiang and Yang, Yujiu},
  booktitle = {International Conference on Learning Representations (ICLR)},
  year      = {2024},
  note      = {arXiv:2309.08532}
}

@inproceedings{meyerson2024lmx,
  title     = {Language Model Crossover: Variation through Few-Shot Prompting},
  author    = {Meyerson, Elliot and Nelson, Mark J. and Bradley, Herbie and Gaier, Adam and Moradi, Arash and Hoover, Amy K. and Lehman, Joel},
  booktitle = {ACM Transactions on Evolutionary Learning and Optimization},
  year      = {2024},
  note      = {arXiv:2302.12170}
}

@inproceedings{castro2026abacus,
  title     = {Abacus: Self-Supervised Event Counting-Aligned Distributional Pretraining for Sequential User Modeling},
  author    = {Castro, Sullivan and Betlei, Artem and Di Martino, Thomas and El Manouzi, Nadir},
  booktitle = {Proceedings of the Nineteenth {ACM} International Conference on Web Search and Data Mining ({WSDM})},
  year      = {2026},
  doi       = {10.1145/3773966.3779391},
  note      = {arXiv:2512.16581}
}

\end{document}